# An automated domain-independent text reading, interpreting and extracting approach for reviewing the scientific literature

**Amauri Jardim de Paula**[*]

Solid-Biological Interface Group (SolBIN),
Department of Physics, Federal University of Ceara – UFC,
P.O. Box 6030, Fortaleza, CE, 60455-900, Brazil
amaurijp@gmail.com

## Abstract

It is presented here a machine learning-based (ML) natural language processing (NLP) approach capable to automatically recognize and extract categorical and numerical parameters from a corpus of articles. The approach (named a.RIX) operates with a concomitant/interchangeable use of ML models such as neuron networks (NNs), latent semantic analysis (LSA), naive-Bayes classifiers (NBC), and a pattern recognition model using regular expression (REGEX). A corpus of 7,873 scientific articles dealing with natural products (NPs) was used to demonstrate the efficiency of the a.RIX engine. The engine automatically extracts categorical and numerical parameters such as (i) the plant species from which active molecules are extracted, (ii) the microorganisms species for which active molecules can act against, and (iii) the values of minimum inhibitory concentration (MIC) against these microorganisms. The parameters are extracted without part-of-speech tagging (POS) and named entity recognition (NER) approaches (i.e. without the need of text annotation), and the models training is performed with unsupervised approaches. In this way, a.RIX can be essentially used on articles from any scientific field. Finally, it can potentially make obsolete the current article reviewing process in some areas, especially those in which machine learning models capture texts structure, text semantics, and latent knowledge.

## 1. Introduction

To extract information in an automated fashion from scientific literature and to convert it into structured databases is a globally recognized data-mining challenge [1], [2]. As documents are written in different formats, with different terminologies, and with a variable level of completeness, there is a certain level of skepticism on the efficiency of information extraction approaches used on current literature [3]. At the same time, information extraction from documents in Chemistry, Physics, and Engineering fields is largely recognized to be an important strategy to fulfill the lack of databases in several areas, especially with information on experimentally tested synthesis protocols, processes, and products. From these perspectives, an automated reading-interpreting-extracting approach (namely a.RIX) is presented here to extract categorical and numerical information from scientific articles and structure it into databases that can be properly analyzed and/or be used for modeling. The a.RIX engine makes use of four trainable modules to localize relevant information in scientific articles: (i) Semantic machine (Sm), (ii) Topic machine (Tm), (iii) Bayesian machine (Bm) and (iv) Section filter (Sf), and a

non-trainable1 (v) Literal Search machine (LSm) that can find terms in plain texts using manually generated REGEX patterns.

Topic-based text classification approaches were already confirmed to be capable of recognizing and classifying inorganic materials synthesis procedures when trained with a corpus of over 2 million scientific articles [4]. However, the efficiency in capturing text structure and text semantics by ML models was demonstrated to be largely correlated with the content of the text corpus used [5]. In other words, a small corpus that is related only to a particular subject can be better captured by ML models than a large corpus with multiple subjects. In this way, we combined multiple ML and NLP approaches in the sense that multiple subject-related categorical and numerical parameters can be extracted even on a small but significantly correlated text corpus. In addition, the engine can be largely optimized considering many possible setups for the Sm, Tm, Bm and Sf, and other models that can be further attached.

When conceiving the a.RIX engine, a certain balance between accuracy and generalization was aimed, especially considering the commonly used NLP pipeline that involves part-of-speech tagging (POS), name entity recognition (NER), and phrase parsing [6]–[11]. As there are big challenges in generalizing POS to multiple domains [8], this process was avoided. Name entity recognition (NER) processes applied concomitantly to Physics, Chemistry, and Engineering fields are also largely impracticable, especially due to huge variability in terminology and the necessity of annotated texts to train NER approaches [4], [12], [13]. Therefore, the use of NER has been avoided because it increases the generalization capacity of the program. As phrase parsing is POS- and NER- dependent, this process was not used as well. Instead, the a.RIX engine was built on machine learning models neuron networks (NNs), latent semantic analysis (LSA), naive-Bayes classifiers (NBC), and on pattern recognition models using regular expression (REGEX). Currently available toolkits such as Chem Data Extractor (which uses POS > NER > parsing approach) are practical to extract information from articles and, in some instances, even to generate structured databases from extracted data [7], [8], [11]. Even though these toolkits have been constantly updated, they perform well only in determined situations (e.g., for specific knowledge domains, articles from specific publishers, and HTML and XML files). On the other hand, a.RIX can be virtually used in any scientific field, and was especially conceived to deal with PDF files, which are ubiquitous among all publishers. To exemplify its capacity, a.RIX processed a relatively small corpus of ~ 7,800 scientific articles related to natural products. The program was able to recognize and extract hundreds of terms related to plants species from which active molecules are extracted, tens of microorganisms against which the extracted active molecules present activity, and the minimum inhibitory concentration (MIC) for these extracted molecules to the microorganisms.

## 2. Methods

### 2.1. Processing the text corpus

As the PDF format is ubiquitous in scientific literature, a focus was given to this format. However, all processes described here that are applied to plain text can be promptly used for articles in HTML and XML formats. In order to perform a suitable conversion of the texts present in the PDF container to a machine-readable plain text format, several text processing filters were implemented in the a.RIX engine using REGEX patterns (especially to standardize physical units, chemical formulas, and others). Considering a large number of publishers from which articles can be downloaded, important text conversion errors occur. After the plain text conversion, the (i) articles sections (e.g., Abstract, Introduction, Methodology, Discussion, Results, References) and the (ii) sentences are identified through REGEX patterns. In the NLP terminology, sentences are used by the a.RIX engine as the basic "document" (or "doc"). In addition, eliminating the References section is imperative to improve the performance of the a.RIX engine, since sentences present in this section are highly uncorrelated. Finally, articles in which the Methodology section was identified are used to train Sf. If well trained, this filter can determine if a particular sentence was extracted from the methodological part of an article, and it is of importance to extract information on synthesis methods and conditions.

## 2.2 Training the a.RIX engine

Sm is trained by clustering semantically similar tokens and n-grams related to relevant "subjects" or semantic relationships. For instance, tokens "wheat", "corn", and "rice" have a large semantic similarity when considering the subject "crop". More specifically, firstly it was obtained a word vector representation for the set of tokens present in the corpus through the Mikolov's word-to-vec (W2Vec) approach [14], [15], which uses a single layered feedforward neural network (FNN) where the number of inputs is set to be the number of unique tokens in the corpus and the neuron number is commonly set to 100-300 (neuron number = D). The word vectors representation (also called word embeddings) is equivalent to the input-neuron weights matrix obtained after the FNN is trained with the whole doc-token one-hot matrix using a modified-continuous bag-of-words (CBOW) method. A word vector is a real-valued vector representation of a token that enables quantitative measurements of similarity. In this way, it was used as input to train Sm a small set of tokens (wheat, corn, and rice) that are possibly related to a relevant subject (crop), and a routine recursively finds tens to hundreds of semantically similar tokens. The semantic similarity between tokens is evaluated by the cosine value between the word vector representation of an input token and all other word vectors (related to the other M − 1 tokens). A large semantic similarity between two tokens means a low cosine value between the word vectors associated with these same tokens. In the first iteration of the machine, the twenty most similar tokens of each input token provided are determined. Further iterations performed by the machine identifies the twenty most similar tokens for each token identified in the previous iteration. After n iterations, the machine counts how many times each token manifested as like any other token. By counting the tokens overlap that occurred in n iterations it is possible to measure of how much a particular token is semantically related to a particular subject. Semantically related tokens identified by Sm can be combined to determine if they appear in the text corpus as n-grams (n > 1). For instance, the token "straw" appears more frequently in the corpus combined with token "wheat" such as "wheat straw". To check if two tokens frequently appear combined in the text as a 2-gram it was used a score function:

$$score(token_i, token_j) = ( counts(token_i, token_j) – thr ) / counts(token_i) . counts(token_j)$$

where *counts(token $_i$, token $_j$)* is a function describing the counts for the 2-gram appearance in the corpus (e.g. "wheat straw"), *counts(token $_i$)* and *counts(token $_j$)* are functions with the counts for each token (e.g., "wheat" and "straw") and *thr* is a arbitrary threshold value.

Tm works with a latent semantic analysis (LSA) model that is trained from the singular value decomposition (SVD) of the transposed TFIDF matrix (TFIDF$^T$). The truncation of the resulting $U_{M,M}$ matrix after SVD can transform a (M x N) token-doc matrix into a (M x E) token-topic matrix, where E is the dimension chosen as the topic vector dimension (usually 100-300). The N x E doc-topic matrix can be obtained from multiplication of TFIDF and token-topic matrix [14]. A relevant topic to be searched in the corpus by Tm can be generated by summing the token-topic vectors of certain tokens related to a determined subject. In this way, it was used as input to train Tm a set of tokens that were previously clustered by subject (e.g. "crop products") with Sm, and the program linearly combines (i.e. sum) the token-topic vectors related to these tokens (token-topic vectors are extracted from the token-topic matrix). The resulting topic vector T generated after combining the token-topic vector is then matched to each doc-topic vector present in the doc-topic matrix. The largest the correlation coefficient calculated for this match, the largest is the probability that a doc (i.e. sentence) correlates with the subject of interest. BOW or one-hot matrices can also be used for training Tm (instead of TFIDF). However, it was confirmed that a lower degree of latent knowledge is captured by the doc-topic and token-topic matrices when one-hot or BOW matrices are used.

Bm is trained with a set of docs (i.e., sentences) that Tm previously selected to a specific subject, i.e., they possess a high correlation coefficient with a determined topic (e.g. "crop products"). A probability mass function (PMF) is obtained for this particular subject by calculating the tokens' appearance in this set of docs. The PMF is used to calculate the probability that other doc is related to the particular subject of interest, i.e., the more the overlap that the tokens appearance distribution of a sentence has with the PMF, the more the probability that the sentence belongs to the particular subject represented by the PMF. Randomly selected sentences with low correlation coefficients for a a subject are used to calculate another PMF to represent a model with a low degree of correlation with the subject of interest. Both PMFs feed a naive Bayes classifier (NBC) that is used to select sentences related to certain relevant subjects.

Sf has an architecture that comprises 1D convolutional neural networks (CNNs) associated with feedforward neuron networks (FNNs). More specifically, inputs go sequentially through two CNNs ($l_1$ and $l_2$) and four FNNs ($l_3$, $l_4$, $l_5$ and $l_6$). A custom setup for the layers is: (i) number of filters for $l_1$ and $l_2$ is 100 and 200, respectively; (ii) kernel size of 3 and 5 for $l_1$ and $l_2$, respectively; (iii) neurons number for $l_3$, $l_4$, $l_5$ and $l_6$ is 500, 100, 10 and 1, respectively; and (iv) layers $l_3$, $l_4$ and $l_5$ have a dropout of 20%. An ELU (exponential linear unit) activation function is used for all layers except for $l_6$, where a single 0/1 output is generated from a sigmoid activation function on its single neuron. All layer weights were initialized with a He normal distribution and a binary cross-entropy loss function was used to evaluate the model precision. A stochastic gradient descent (SGD) is used as the training optimizer. Sf is trained using a set of paragraphs automatically extracted and labeled during the text processing step. Each paragraph contains a set of docs (sentences) labeled in regard to the article section from which they were extracted (Introduction, Experimental Methods, Discussion, Conclusions, References).Sf is trained mainly to be used as a classifier to determine whether a relevant doc (sentence) is part of the Experimental part of the article. A detailed diagram showing the text processing and training steps is represented in Figures 1, 2 and 3, and described in Table 1.

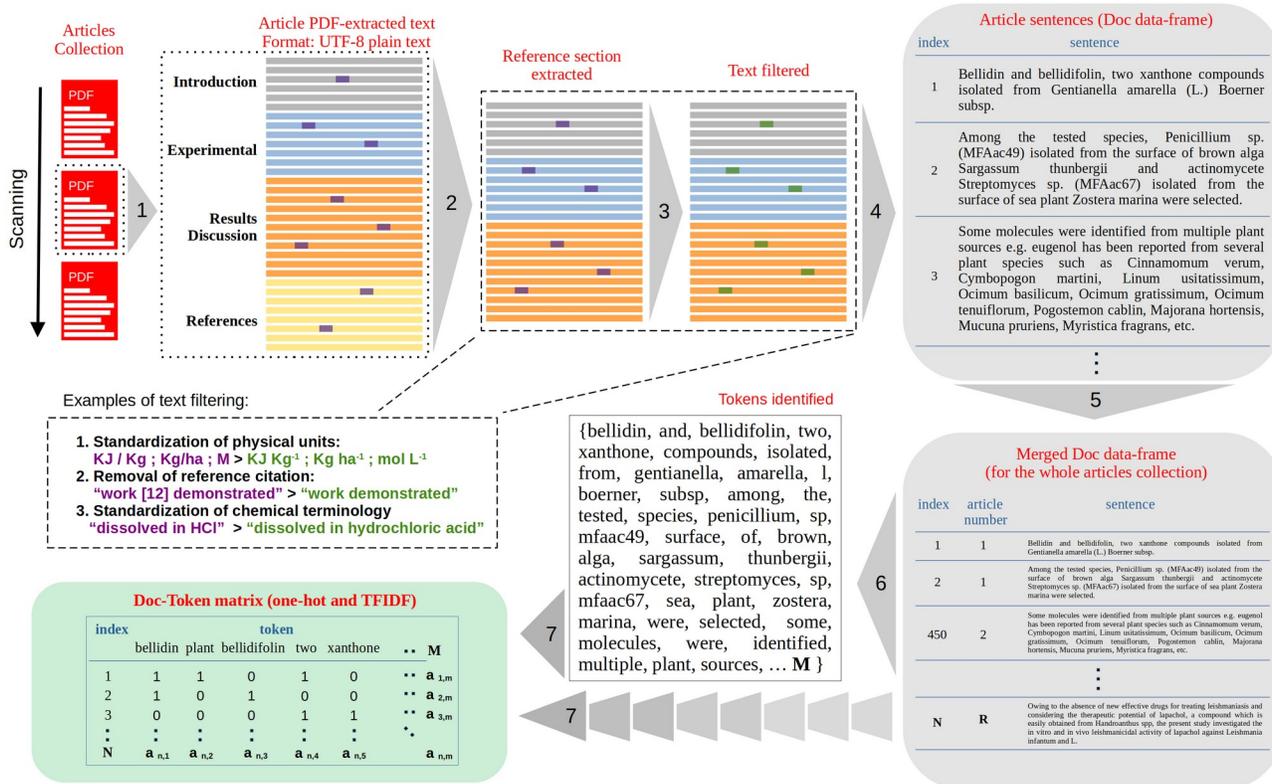

**Figure 1.** Diagram representing data preparation steps: PDF conversion to UTF-8 plain text; article section labeling and extraction (References elimination); text filtering; text breaking into sentences (organized as data-frames); data-frame merging (including all articles in the collection); tokens identification; and generation of two doc-token matrices (one-hot and TFIDF). In this figure, the doc-token matrix with one-hot vectors is depicted (green box). Details on processes 1 to 7 are described in Table 1.

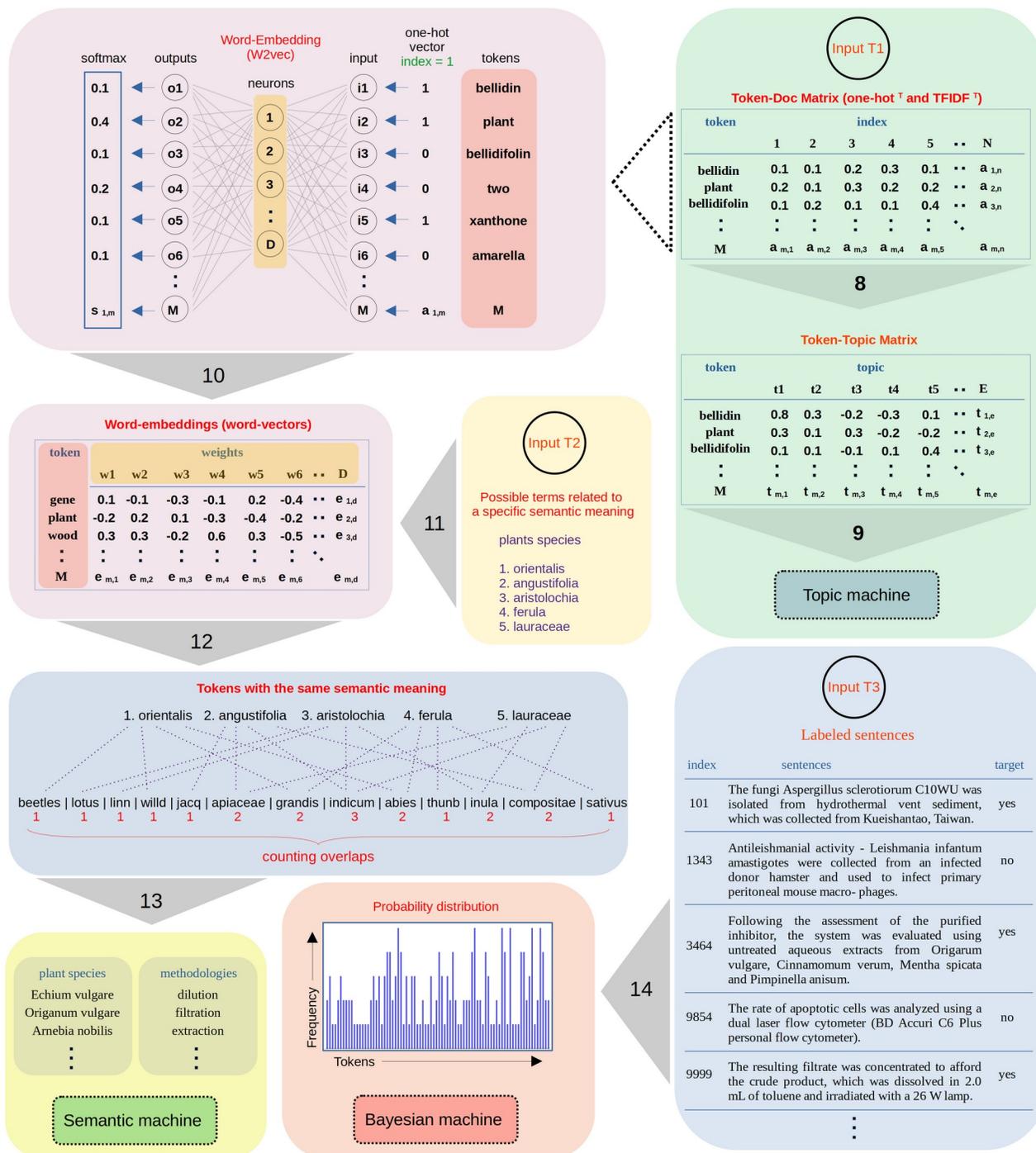

**Figure 2.** Diagram showing how Semantic, Topic and Bayesian machines (Sm, Tm and Bm, respectively) are trained. Tm is obtained from the singular value decomposition (SVD) of transposed doc-token matrices (one-hot$^T$ and TFIDF$^T$), known as token-doc matrices (Input T1). Better results are achieved using the TFIDF$^T$ (top matrix in the green box). Sm is obtained by finding the similarity of word vectors (measured by cosine) related to relevant subjects (Input T2). Bm is trained with labeled sentences that carry relevant information (Input T3). Details on processes 8 to 14 are described in Table 1.

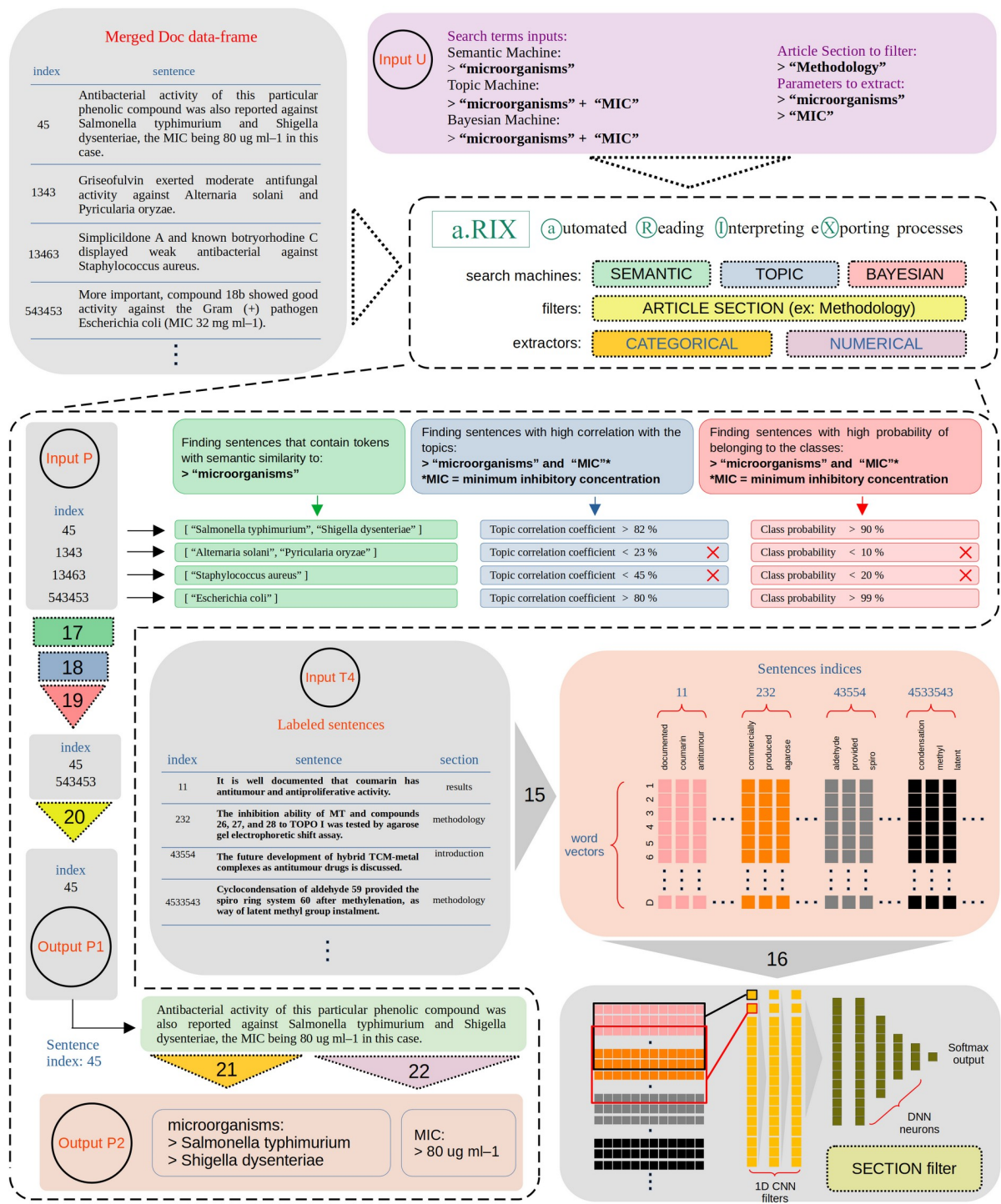

**Figure 3.** Diagram depicting the training process of the Section filter (Sf) through section-labeled sentences (Input T4), and depicting the processes performed by a full-trained a-RIX engine. The engine works with inputs from the user (Input U) and from the articles collection (Input P). The engine performs extraction of relevant sentences (Output P1) and categorical/numerical parameters (Output P2). Details on processes 15 to 22 are described in Table 1.

**Table 1.** Description of process related to data preparation and training of Sm, Tm, Bm and Sf and operation of the *a-RIX* engine.

| Process number | Input | Output | Description |
|---|---|---|---|
| 1 | • PDF file | • article plain text | Conversion of the PDF text to UTF-8 plain text using freely available PDF text tool-kits [6], and labeling of articles sections such as Introduction, Experimental part, Results, Discussion and References. |
| 2 | • article plain text | • article plain text (filtered) | Removal of the References section. |
| 3 | • article plain text (filtered) | • article plain text (filtered) | Text filtering using REGEX patterns for standardization of the text and correction of text conversion issues. |
| 4 | • article plain text (filtered) | • doc data-frame | Text breaking with REGEX patterns and data-frame conversion using Pandas [16]. To adopt a NLP common terminology, the article sentence is the **"document"** (i.e., **doc**) upon which basic processing is performed. |
| 5 | • doc data-frames | • doc data-frame (merged) | Merging of data-frames obtained for each article in the collection. The total number of documents (sentences) for the whole collection of scientific articles is defined by **N** and the number of scientific articles is defined by **R**. |
| 6 | • doc data-frame (merged) | • tokens list | Identifying unique tokens in the whole collection (considering all scientific articles). The number of tokens is defined by **M**. |
| 7 | • doc data-frame (merged)<br>• tokens list | • doc-token matrix (one-hot and TFIDF) | Generation of document-token **(doc-token)** matrices. Two doc-token matrices are generated, (i) the first with **one-hot** vectors and (ii) the second with term frequency–inverse document frequency (**TFIDF**) vectors [15]. |
| 8 | • token-doc matrix (TFIDF $^T$) | • token-topic matrix<br>• doc-topic matrix | Generation of **token-topic** matrices by Singular Value Decomposition (**SVD**) of the transposed TFIDF matrix (TFIDF $^T$), named **token-doc** matrix (**Input T1**). SVD results in: $$TFIDF_{N,M}^T = U_{M,M} \cdot S_{M,N} \cdot V_{N,N}^T$$ where U contains left singular vectors, S contains singular values and $V^T$ contains right singular vectors. The topic vector dimension is defined by the value **E** (t1, t2, t3, ..., E), which is determined during the truncation of $U_{M,M}$, when only the first E columns (from left-right) are collected, thus resulting in a $U_{M,E}$ matrix. The token-topic matrices are equivalent to: $$token\text{-}topic_{M,E} = truncated\ (U_{M,M}) = U_{M,E}$$ and contain the **token-topic vectors**. The doc-topic matrices are obtained by: $$doc\text{-}topic_{N,E} = TFIDF_{N,M} \cdot U_{M,E}$$ This method is known by Latent Semantic Analysis (**LSA**) [15]. |
| 9 | • token-topic vectors | • topic vector | A topic vector that carries information on a determined subject is obtained by summing token-topic vectors associated with tokens related to that subject. A collection of relevant topic vectors feeds the **Topic Machine (Tm)**. |
| 10 | • doc-token matrix (one-hot) | • token-weights matrix | Word embeddings is obtained through Word-to-vector (W2vec) approach using feedforward neuron networks (FNNs) [16]. The whole doc-token matrix (with one-hot vectors) is used to generate the word embeddings. A modified-continuous bag of words (CBOW) approach is used to sequentially train the FNN [15]. Each sentence is split in small chunks (3-6 tokens) and introduced in the FNN as one-hot vectors. This approach is performed for the FNN to capture both short- and long-range correlations between the tokens present in sentences. After the FNN training, the **word vectors** for this method correspond to the resulting neurons weights associated with all M tokens. The dimension of the word vectors is defined by the number of neurons in FNN (**D**). The neuron network architecture was constructed on Keras [17]. |
| 11 | • list of tokens | • word vectors | Selection of word vectors associated with the tokens introduced by the user (**Input T2**). It is assumed that the tokens introduced are related to a relevant subject. |
| 12 | • word vectors | • list of tokens | Search among the M tokens of the collection that have semantic similarity with those introduced by the user (Input T2). Determination of semantic similarity is performed by measuring the cosine between the word vectors associated with the tokens introduced by the user (Input T2) and other M tokens. |
| 13 | • list of tokens | • tokens data-frame | Organization of semantically similar tokens into data-frames. The **Semantic Machine (Sm)** contains sets of tokens related to relevant subjects. |
| 14 | • doc data-frame (labeled) | • Probability distributions | Determination of probability distributions for relevant subjects. Probability distributions are calculated by summing and normalizing bag-of-words (**BOW**) vectors determined for labeled sentences (**Input T3**). The **Bayesian machine (Bm)** contains sets of probability distributions related to relevant subjects. |
| 15 | • doc data-frame (labeled) | • word vectors (concatenated) | Conversion of labeled sentences (**Input T4**) to stacked word vectors and further concatenation of stacked word vectors from multiple sentences related to a determined article section, such as the Experimental part (i.e., Methodology). The concatenation of word vectors has the effect of putting together sentences (similar to a paragraph) of a similar article section. |
| 16 | • word vectors (concatenated) | • CNN model | Training of the 1D convolutional neural network (CNN) from concatenated word vectors assembled from multiple sentences. The **Section filter (Sf)** comprises the trained 1D CNN. The CNN architecture was generated with Keras [17]. |
| 17 | • doc data-frame (merged)<br>• User inputs | • Filtered sentences | **Sm** selects only documents (sentences) that contain tokens related to a determined subject. |
| 18 | • Filtered sentences | • Filtered sentences | **Tm** selects only documents (sentences) for which the topic vectors possess a high correlation (evaluated with the correlation coefficient) with the topic vector of a determined subject. |
| 19 | • Filtered sentences | • Filtered sentences | **Bm** selects only documents (sentences) for which the normalized BOW vectors match with the probability distribution calculated for a determined subject. |
| 20 | • Filtered sentences | • Filtered sentences | **Sf** selects only documents (sentences) that are related to a determined article section. |
| 21 | • Filtered sentences | • Categorical parameter | Extraction of **categorical** parameters from selected sentences. |
| 22 | • Filtered sentences | • Numerical parameter | Extraction of **numerical** parameters from selected sentences. |

## 2.3 Calculating the recall and accuracy

The a.RIX engine recall for any categorical or numerical parameter extracted was calculated by:

$$recall_R(parameter) = ne_R(parameter) / R$$

where $ne_R(parameter)$ is the number of articles from which a determined categorical or numerical parameter was extracted and $R$ is the total number of articles in the corpus. However, to accurately determine if a particular categorical or numerical parameter is present in all R articles is impracticable even for a small corpus with tens of thousands of articles. In this way, the recall value is underestimated when using the total number of articles (R) in the equation for $recall_R$. Therefore, another estimation for the recall ($recall_{R50}$) was conceived by randomly sampling 50 articles in the corpus and manually checking each article to determine whether a parameter was present in the article. The $recall_{R50}$ and $accuracy_{R50}$ were then calculated by:

$$recall_{R50}(parameter) = ne_{R50}(parameter) / R50_{parameter}$$
$$accuracy_{R50}(parameter) = nc_{R50}(parameter) / R50_{parameter}$$

where $R50_{parameter}$ represents 50 randomly chosen articles in which a determined parameter was confirmed to be present (thus $R50_{parameter} = 50$), $ne_{R50}(parameter)$ is the number of articles (among $R50_{parameter}$) from which the program extracted any information, and $nc_{R50}(parameter)$ is the number of articles (among $R50_{parameter}$) from which the parameter was correctly extracted.

## 2.4. Extracting categorical and numerical parameters

For categorical parameters, the Categorical extractor (CE) gets from selected sentences the n-grams that Sm previously grouped using REGEX patterns. The n-grams that appeared both in the singular and plural form can be automatically grouped through routines (e.g., "olives" is grouped with "olive") while n-grams with a similar meaning can be manually grouped (e.g., "maize" is grouped with "corn"). On the other hand, numerical parameters can be extracted with the Numerical extractor (NE), which was also built on REGEX patterns. The recognition of a relevant physical or chemical unit (e.g., °C, g mL$^{-1}$, mol L$^{-1}$) in the sentence is useful for the extraction of numerical parameters. In this sense, NE scans a sentence looking matches for a huge vast variety of physical and chemical parameters described in hundreds of units of measurements (e.g., mg mL$^{-1}$, mJ, m$^2$ g$^{-1}$). In addition, dimensionless parameters (e.g., percentage, ratio between physical units, to name a few) can also be extracted from sentences.

NE works in two extraction modes: (i) All-Val (AV) and (ii) Grouped-Val (GV). AV extraction is suitable when one seeks to survey just one numerical parameter per document. However, if an association between multiple numerical parameters is aimed, especially for building high dimensional databases (with multiple features), AV extraction largely reduces both the recall and accuracy of the resulting database (see Table 2). For instance, in a first scan performed to extract the parameter synthesis temperate, the NE operating on AV mode would extract from the sentence "The material X was synthesized at 300, 400, 500, 600, and 700 °C, and samples were named X300, X400, X500, X600 and X700" the values 300, 400, 500, 600 and 700 (number of extracted values $n_1$ = 5 values) . These values are indexed such as $i_1\{300\}$, $i_2\{400\}$, $i_3\{500\}$, $i_4\{600\}$ and $i_5\{700\}$ and organized into a data frame. In a second scan performed to find the parameter surface area, in AV mode, NE will only extract values from a sentence for which the values extracted ($n_2$) match the number of values extracted in the first scan ($n_2$ must be equal to $n_1$). Thus, if the sentence "Samples X400, X500 and X600 have surface area values of 345, 103, and 149 m$^2$ g$^{-1}$, while measurements for samples X300 and X700 could not be performed" is processed by NE (in AV mode) in the second scan, no values of surface area are extracted because $n_2$ = 3 and $n_1$ = 5 (see Table 2).

In a situation that $n_2$ = $n_1$, e.g. "Samples X400, X500 and X600 have surface area values of 345, 103, and 149 m$^2$ g$^{-1}$, while X300 and X700 were 45 and 53 m$^2$ g$^{-1}$", NE (in AV mode) extracts and indexes surface area values according to the previous indexes set for synthesis temperature, by following the order of appearance in the sentence, i.e., $i_1\{300, 345\}$, $i_2\{400, 103\}$, $i_3\{500, 149\}$, $i_4\{600, 45\}$ and $i_5\{700,53\}$ (see Table 2). As a result, the resulting association between synthesis temperature and surface area can be inaccurate. To ensure accuracy when extracting multiple parameters in AV mode, NE can be set to extract values just from sentences in which a single numerical value is present (e.g., "The material X was synthesized at 700 °C" and "The surface area of the material X was of 56 m$^2$ g$^{-1}$"), though the recall is largely affected with this setup (see Table 2). Another approach for getting an accurate association between multiple numerical parameters during the extraction process can be conceived through a name entity recognition (NER) method through which sample names are identified when first mentioned in the text. These entities recognized as samples in the articles can be further used to localize the associated numerical parameters related to each of them. However, due to the lack of standardization in the terminology used for sample description among articles, this approach becomes largely impracticable even using a huge amount of annotated texts.

On the other hand, when NE operates in GV mode, the program groups all numerical values related to a parameter and a statistical representation of these values are used. In this way, for each numerical parameter extracted from each article, one can get an average (*avg*) value, an interval (*int*) value, a lowest (*l*) value and a highest (*h*) value. Other relevant statistical representations of the grouped values (e.g., median and standard deviation) can also be used if $n_i > 10$. Advantages of the GV mode are increased recall when multidimensional databases are aimed and a suitable article-to-article statistical comparison for multiple numerical parameters. However, when the database will be used for predictions, the AV mode is preferable.

**Table 2.** Operation modes for the Numerical extractor

| NE mode | 1° scan | | 2° scan | |
|---|---|---|---|---|
| | sentence to extract | Numerical parameter to extract: synthesis temperature (°C) | sentence to extract | Numerical parameter to extract: surface area (m² g⁻¹) |
| all values (AV) | The material X was synthesized at **300, 400, 500, 600, and 700 °C**, and samples were named X300, X400, X500, X600 and X700 | number of single values extracted ($n_1$): 5 — values extracted:[1] $i_1$:{300} $i_2$:{400} $i_3$:{500} $i_4$:{600} $i_5$:{700} | Samples X400, X500 and X600 have surface area values of **345, 103, and 149 m2 g⁻¹**, while measurements for samples X300 and X700 could not be performed | number of single values extracted ($n_2$): 3 — cumulative values extracted ($n_2 \neq n_1$):[2] $i_1$:{300, None} $i_2$:{400, None} $i_3$:{500, None} $i_4$:{600, None} $i_5$:{700, None} |
| | | | Samples X400, X500 and X600 have surface area values of **345, 103, and 149 m2 g⁻¹**, while X300 and X700 were **45 and 53 m2 g⁻¹** | number of single values extracted ($n_2$): 5 — cumulative values extracted ($n_2 = n_1$):[3] $i_1$:{300, 345} $i_2$:{400, 103} $i_3$:{500, 149} $i_4$:{600, 45} $i_5$:{700, 53} |
| grouped values (GV) | | number of grouped values extracted ($n_1$): 1 — values extracted: $i_1$:{300–700} | Samples X400, X500 and X600 have surface area values of **345, 103, and 149 m2 g⁻¹**, while X300 and X700 were **45 and 53 m2 g⁻¹** | number of grouped values extracted ($n_2$): 1 — cumulative values extracted ($n_2 = n_1$):[4] $i_1$:{300–700, 45–345} |
| all values (AV) | The material X was synthesized at **700 °C** | number of grouped values extracted ($n_1$): 1 — values extracted: $i_1$:{700} | The surface area of the material X was of **56 m2 g⁻¹** | number of grouped values extracted ($n_2$): 1 — cumulative values extracted ($n_2 = n_1$):[5] $i_1$:{700, 56} |

[1] Values extracted in the first NE scan are indexed following their order of appearance in the sentence.

[2] As there are $n_1 = 5$ values extracted in the 1° scan (to get the parameter 'synthesis temperature'), during the 2° scan (to get the parameter 'surface area') the extractor will try to get $n_2 = 5$ values. If $n_2 \neq n_1$, the extraction process is canceled and a None value is combined to each of the indexed values.

[3] As there are $n_1 = 5$ values extracted in the 1° scan and $n_2 = 5$ values extracted during the 2° scan (thus $n_1 = n_2$), NE extracts and combines the values extracted in the 2° scan by following their order of appearance in the sentence.

[4] In GV mode, NE extracts parameters with numerical values by grouping all values in a unique index, and statistical representations of these grouped values can be used such as the average value, the highest value, the lowest value, and others.

[5] In order to increase the accuracy when assembling multidimensional databases (with multiple features), NE can be set to extract a numerical parameter just from sentences in which a single numerical value is present.

## 3. Experiments

Possible 2-grams related to plant species were identified through the Semantic machine (Sm). As an input to this machine, we used a small set of tokens such as *"rubiaceae"*, *"acacia"*, *"sophora"*, *"pinaceae"*, *"prunus"*, *"polygonaceae"*, and *"japonica"*, which were identified by manually opening some random articles of the collection. In this way we were able to recognize hundreds of 2-grams related to the subject "plant species" (see Table S1). In most cases of 2-grams found by Sm, it was possible to capture and group both tokens related to the genus and species. Common plant genus and species found are depicted in Fig 4. The identification of categories related to the kingdom, phylum, class, order and family was performed by comparing these 2-grams in the Global Biodiversity Information Facility (www.gbif.org). In this way, we observed that 2-grams related to other species (besides plants) could also determine by Sm when plants genus and species were used as input. In addition, Sm was also used to find tokens related to subjects "microbial species" and "experimental methods used in articles". For these subjects, the 2-gram representation is accurate to describe microorganisms species (e.g. *Enterobacter aerogenes*, *Fusarium solani*, and *Klebsiella pneumonia*) and 1-gram representation is accurate to describe experimental methods (e.g. sonicated, homogenized, incubated, and agitated).

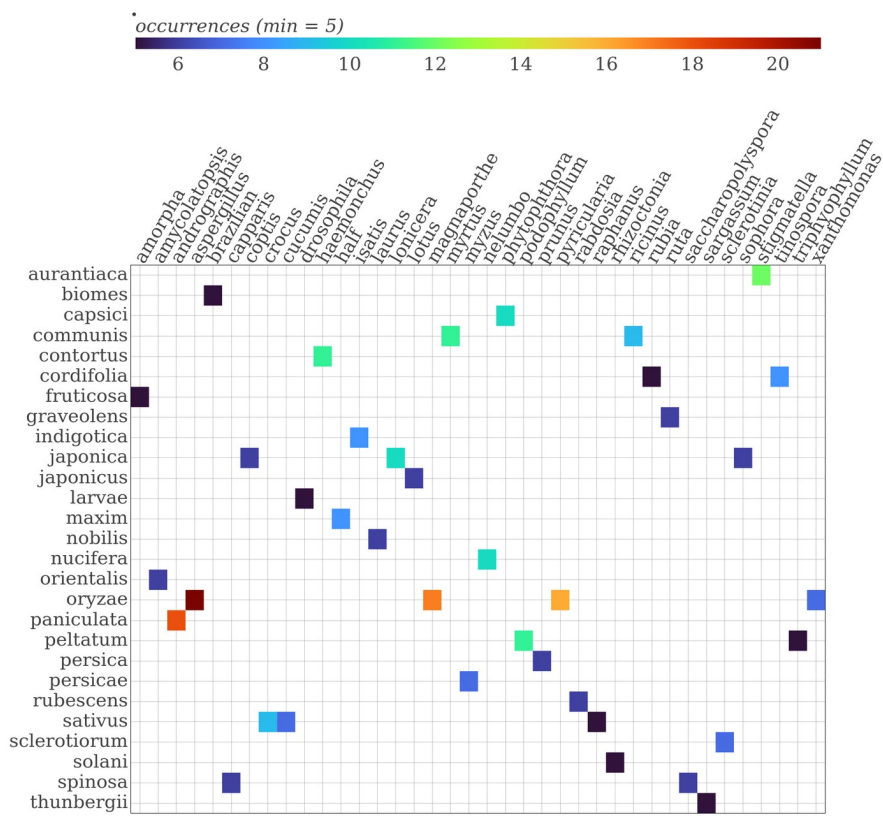

**Figure 4.** Gridplot showing tokens correlation used to determine the plant species described in each article. Genus is represented in X-axis while species is represented in Y-axis. Only terms that were identified in at least 5 articles are represented.

The Topic machine (Tm) then generates a topic vector $T_i$ related to each subject for which Sm previously grouped the tokens. $T_i$ is generated by summing the token-topic vectors representation of the tokens belonging to each subject. This approach is known as latent semantic analysis (LSA). "Plants species" and "experimental methods" are subjects that were represented by two topic vectors: $T_1$ and $T_2$, respectively. It is also possible to sum $T_1$ and $T_2$ to obtain a new topic related to the "experimental methods applied on plant species". The resulting $T_1+T_2$ topic was used to localize in the corpus the sentences that accurately describes the plants species from which possible active molecules or extracts were isolated or purified. If only $T_1$ matches are searched, sentences that just mention plants species can be retrieved, such as: "Similar to *Artemisia diffusa*". Instead, if $T_1$ is combined with $T_2$, the engine will search for sentences like: "Luteolin and luteolin–7methylether were isolated from the Thai plant *Coleus parvifolius* Benth". The Section filter (Sf) can be used to confirm that such sentences are from the Experimental part of the articles. Finally, the Literal Search machine (LSm) can be used when a specific term or combinations of terms must be found in the sentence. When looking for sentences describing active molecules or extracts isolated from plants, LSm can be set with the input "[isolat*, extract*]". In this way, tokens such as 'isolated' or 'extraction' must be present in the sentence for the match to occur. In another situation, if it is aimed to find the minimum inhibitory concentration (MIC) of an isolated molecule against microorganisms, the input for LSm should be "[n{1,4}{conc}]". With this input, the program will search for sentences containing any numerical value between "1" and "9999" combined with a concentration unit (e.g., g $L^{-1}$, mg $L^{-1}$, mol $L^{-1}$, mmol $L^{-1}$).

In order to evaluate the a.RIX engine recall and accuracy when dealing with the NPs database, several search-extract routines were performed using different setups in the program to to find terms (2-grams) related to plants species (see Table 3). Overall, by combining multiple search inputs in LSm and Bm it is possible to increase the values of *accuracy*$_{R50}$ along with a reduction in the values of *recall*$_{R50}$ (compare setups 1 and 2 in Table 3). The topic overlap can also increase the values of *accuracy*$_{R50}$, especially if information present in a more particular context is aimed. For instance, when it is aimed to extract only the plant species from which active molecules were extracted, it is more appropriate to use a topic combination in Tm (see setup 3 in Table 3). When searching for information present in the Experimental part of articles, Sf is recommended since it can increase the *accuracy*$_{R50}$ (compare setups 4 and 5 in Table 3). However, CNN filters such as Sf requires a higher computational cost. Finally, by tuning the threshold for the Tm correlation coefficient, it is possible to largely change the *recall*$_{R50}$ and *accuracy*$_{R50}$ values (compare setups 6 and 7 in Table 3).

REGEX can also improve the values of *accuracy*$_{R50}$ when information to be extracted is associated in sentences with known characters patterns. For example, to extract the terms (2-grams) describing microbial species for which minimum inhibitory concentration (MIC) was calculated, it is possible to construct a REGEX for finding numerical characters associated with concentration units in sentences. Both the microbial species (see setup 8 and 9 in Table 3) and MIC values (see setup 10 and 11 in Table 3) can be extracted with the same setup. When assembling multi-parameter databases through the use of multiple scans on the a.RIX engine, both microorganisms species and MIC values can be associated to plant species used to extract the active molecules (see an example in Table 2).

**Table 3.** Recall and accuracy of a.RIX engine as a function of the setup (using NPs database)

| Information Searched | Setup | Inputs for searching in the corpus | | | | | Extractors | | Recall / Accuracy | | |
|---|---|---|---|---|---|---|---|---|---|---|---|
| | | LSm | Tm | Tm corrcoef threshold[1] | Bm | Sf[2] | Categorical | Numerical | $recall_R$ (%) | $recall_{R50}$ (%) | $accuracy_{R50}$ (%) |
| plant species | 1 | plants | None | None | None | None | plants | None | 19 | 40 | 67 |
| | 2 | plants | plants | 0.5 | None | None | plants | None | 7 | 15 | 81 |
| | 3 | plants | plants + processes | 0.5 | None | None | plants | None | 8 | 15 | 86 |
| | 4 | plants + [isolat*, extract*] | None | None | None | None | plants | None | 8 | 14 | 92 |
| | 5 | plants + [isolat*, extract*] | None | None | None | on | plants | None | 7 | 12 | 95 |
| | 6 | None | plants + processes | 0.6 | None | None | plants | None | 53 | 66 | 88 |
| | 7 | None | plants + processes | 0.7 | None | None | plants | None | 7 | 14 | 93 |
| microorganisms | 8 | [n{1,4}{conc}] | None | None | None | None | microorganism | None | 2 | 51 | 90 |
| | 9 | [n{1,4}{conc}] | microorganisms | 0.5 | None | None | microorganism | None | 1 | 42 | 95 |
| MIC | 10 | [n{1,4}{conc}] | None | None | None | None | None | conc | 2 | 52 | 97 |
| | 11 | [n{1,4}{conc}] | microorganisms | 0.5 | None | None | None | conc | 1 | 40 | 98 |
| processes | 12 | processes | processes | 0.7 | None | on | processes | None | 35 | 42 | 96 |
| | 13 | processes | None | None | processes | on | processes | None | 75 | 76 | 97 |

[1] In Tm, a sentence is related to a determined subject if the doc-topic vector representation of the sentence and the topic vector representing the subject have a correlation coefficient (corrcoef) value larger than the threshold. A corrcoef of 1.0 represents 100% of correlation between two vectors.

[2] Sf was trained to determine if a particular sentence is a part of the Experimental part of articles

The Bayesian machine (Bm) represents an approach to generalize the information extraction model for new articles that were not used to generate the doc-token matrices, i.e., new articles added to the collection after the Sm, Tm, and Sf were trained. To obtain a PMF related to a determined topic that will be used in Bm, it is recommended to increase the Tm correlation coefficient in order to extract a set of sentences with a high value of $accuracy_{R50}$. This approach generalizes the model to extract sentences related to experimental processes described in articles (see setup 12 in Table 3). The Bm accuracy towards new articles can also be improved by Sf (see setup 13 in Table 3).

## 4. Discussion

Sm and Tm essentially capture text semantics. In the case of Sm, identifying the semantic context in which a particular term (n-gram) is used depends on the availability of sentence structures. More specifically, to identify that two n-grams share the same meaning in a determined context, the words embeddings (W2Vec) present in Sm must be trained from several sentences in which the two n-grams share similar neighbor tokens. For instance, consider sentences "...leaves of *Abies grandis* were used to extract molecule X..." and "...molecule Y has a high antimalarial activity and it is obtained from *Dipteryx odorata* leaves...". If many sentences with such structure are available in the corpus for terms *Abies grandis* and *Dipteryx odorata*, Sm will cluster these 2-grams as similar. In other words, the presence of several sentences in which a term is used with a similar meaning is imperative for Sm to capture semantics. On the other hand, the sentence variability should not be large enough to result in multiple meanings for a determined n-gram. This is the reason why a.RIX engine performs well with a corpus of a specific scientific field, for which n-grams present a unique and defined meaning.

Regarding Tm, the accurate definition of a particular topic also depends on a large availability of sentences with structures defined by a particular n-gram distribution. The engine can be tuned by combining topic vectors when specific categorical parameters are aimed. For instance, when it aims to extract just the plant species used in articles, a single topic vector can be used based on the set of 2-grams clustered by Sm. In addition, when it aims to extract plant species processed to extract a particular active molecule, a topic vector combination could perform with the set of tokens related to the plant species and the set related to experimental processes (both clustered with Sm). Further topic vector combinations (*n* topics > 2) can be used to increase the engine accuracy values.

A suitable setup for Tm can also be used to extract numerical parameters such as the minimum inhibitory concentration (MIC), i.e., a numerical parameter given in a unit of mass per volume, which is commonly associated with a categorical parameter, i.e., the microorganism species for which the MIC value was assessed. In this context, firstly, a topic vector can be assembled by Tm with the tokens related to microorganisms that Sm clustered. In addition, LSm can be set to find REGEX patterns related to concentration (ex: mg $L^{-1}$). With this concomitant use of a.RIX engine machines, the accuracy is largely increased.

For certain corpus of scientific articles, the use of the a.RIX engine can completely replace the manual article reviewing process currently used to acquire an updated knowledge of a specific scientific field. The engine performs outstandingly well in corpus in which there is a vast variety of categorical parameters (e.g., living species, raw materials, and reagents) associated with numerical parameters in sentences. This occurs because most numerical parameters are represented with standardized physical units (e.g., distance, time, energy, area, volume, and concentration). In addition, for topics presented with standardized text structures, including the use of very specific tokens, the engine can achieve a good generalization towards new articles, which are not present in the processed corpus. Finally, the a.RIX engine can assemble databases with multiple parameters, both categorical and numerical, which can be used for further modeling of multiparametric phenomena.

## 5. Acknowledgments

Finance support was granted by Fundação Cearense de Apoio ao Desenvolvimento Científico e Tecnológico (FUNCAP, Grants PRONEX PR2-0101-00006.01.00/15 and PRONEM PNE-0112-000480100/16) and National Council for Scientific and Technological Development (CNPq, Grants numbers 423567/2018-7 and 308047/2018-4).

# A. Supplementary Information

**Table S1.** Subject-related sets of n-grams grouped by Sm

| subject | Classification | Set of n-grams[1] |
|---|---|---|
| plants species | 2-gram | Abies grandis, Abutilon indicum, Acanthamoeba trophozoites, Achyranthes japonica, Aciculitis orientalis, Acnistus arborescens, Acoelomorpha flatworms, Acremonium pers, Acremonium strictum, Actaea racemosa, Aedes aegypti, Aedes albopictus, African lianas, African pers, Agathis macrophylla, Aglaia odorata, Aglaia roxb, Aglaonema hook, Albourz mountains, Alisma orientalis, Alkanna orientalis, Alkekengi officinarum, Alnus japonica, Alnus sieb, Aloe arborescens, Alpinia officinarum, Alstonia angustifolia, Alstonia macrophylla, Alternaria brassica, Alternaria solani, Amorpha fruticosa, Amphidinium dinoflagellates, Amycolatopsis orientalis, Amycolotopsis orientalis, Ancistrocladus lianas, Andrographis paniculata, Andrographis paniculata, Anethum graveolens, Anoectochilus roxb, Anopheles stephensi, Anopheles subpictus, Anthemis nobilis, Apium graveolens, Aplysia larvae, Apostichopus japonicus, Aptenia cordifolia, Aralia cordata, Araucarea angustifolia, Araucaria angustifolia, Arctoscopus japonicus, Ardisia japonica, Argania spinosa, Arnebia nobilis, Artemia nauplii, Artemisia arborescens, Artemisia diffusa, Arthrospira maxim, Artocarpus communis, Artocarpus nobilis, Asobara pers, Aspergillus japonica, Aspergillus japonicus, Aspergillus oryzae, Aucklandia lappa, Austroeupatorium inula, Babylonia japonica, Balanus amphitrite, Balanus larvae, Balsamocitrus paniculata, Barringtonia racemosa, Bauhinia racemosa, Beltraniella japonica, Berberis microphylla, Berberis thunb, Biomphalaria snails, Biota orientalis, Bipolaris oryzae, Boerhaavia diffusa, Boerhavia diffusa, Bombina maxim, Bonara tomentosa, Boreava orientalis, Botrylloides pers, Botulinum neurotoxins, Bracon wasps, Breynia fruticosa, Brunsvigia orientalis, Bulinus snails, Bupleurum boiss, Byrsonima microphylla, Caladenia orchid, Callicarpa americana, Callicarpa japonica, Callicarpa macrophylla, Calotropis procera, Cambodian provinces, Camellia japonica, Cameroonian rutaceae, Camponotus japonicus, Cananga odorata, Cannabis sativus, Caparis spinosa, Capparis spinosa, Capsicum annuum, Capsicum frutescens, Caragana arborescens, Caragana microphylla, Cardamine cordifolia, Caribbean gorgonians, Carpinus cordata, Carrion beetle, Casearia lucida, Cassia angustifolia, Cassia roxb, Cassia sieb, Caulerpa racemosa, Cedrela odorata, Cellvibrio japonica, Centaurea spinosa, Centaurium erythraea, Cephalosporium acremonium, Ceratocystis picea, Ceroglossus beetle, Cerrado biomes, Chiloglottis orchid, Chirang districts, Chironomus larvae, Chloranthus japonicus, Chromolaena odorata, Chrysanthemum indicum, Cimicifuga racemosa, Cinnamomum zeylanicum, Citrus maxim, Citrus nobilis, Coccinia grandis, Cocos nucifera, Codiaeum peltatum, Coleus blume, Colobometra pers, Colvillea racemosa, Cone snails, Conus snails, Coptis japonica, Cordyceps indigotica, Cossidae moths, Cratoxylum arborescens, Crocus sativus, Cryptomeria japonica, Cucumis sativus, Cucurbita maxim, Culex larvae, Culex mosquito, Culex pipiens, Culex quinquefasciatus, Cunonia macrophylla, Curcurbita maxim, Cyclamen pers, Cytospora eugenia, Dalbergia paniculata, Dalea spinosa, Dead larvae, Dendranthema indicum, Desfontainia spinosa, Diabrotica beetle, Dianthus japonicus, Dicerocaryum senecio, Dicoma tomentosa, Dictyota men, Dioncophyllaceae lianas, Diospyros lotus, Dipteryx odorata, Dittrichia graveolens, Dracaena angustifolia, Drimia robusta, Drosophila flies, Drosophila larvae, Drosophilla flies, Echinacea angustifolia, Echinogorgia aurantiaca, Echinophora spinosa, Echium invaders, Ecklonia maxim, Eimeria maxim, Eleagnus angustifolia, Endlicheria paniculata, Engelhardia roxb, Eplingiella fruticosa, Eriobotrya japonica, Eucalyptus robusta, Euphorbia officinarum, Excoecaria lucida, Fagaceae asteraceae, Fagopyrum esculentum, Fagraea blume, Fallopia japonica, Femto maxim, Feroniella lucida, Ferula communis, Ferula orientalis, Ferula pers, Ficus racemosa, Fissidens nobilis, Flemingia |

macrophylla, Flos sophora, Foeniculum vulgare, Fraxinus angustifolia, Fritillaria thunb, Fusarium solani, Galeopsis angustifolia, Garcinia macrophylla, Garea macrophylla, Gastrointestinal nematode, Gentiana macrophylla, Geodia japonica, Geodinella robusta, Gersemia fruticosa, Giardia cysts, Giardia trophozoites, Glossodoris molluscs, Glossophora kunth, Glycyrrhiza foetida, Haemonchus contortus, Halichondria japonica, Hedyotis diffusa, Helianthus annuus, Heliotropium indicum, Helminthosporium oryzae, Hemonchus contortus, Heracleum maxim, Heteromorpha arborescens, Heterosiphonia japonica, Himalayan mountains, Hippocampus japonicus, Hippospongia communis, Holothuria nobilis, Hordeum vulgare, Hortonia angustifolia, Houttuynia cordata, Humulus japonicus, Hydrangea macrophylla, Hygrophoropsis aurantiaca, Hypnea japonica, Hyptis fruticosa, Ilyonectria robusta, Inula hook, Inula japonica, Inula racemosa, Isatis indigotica, Isodon japonicus, Isodon rubescens, Isoria japonica, Jatropha curcas, Juniperus communis, Jurkat ells, Kadsura angustifolia, Keiskea japonica, Kopsia fruticosa, Laguncularia racemosa, Laminaria japonica, Launaea arborescens, Laurus nobilis, Lavandula angustifolia, Lawsonia spinosa, Leea macrophylla, Leonurus japonicus, Ligularia macrophylla, Ligusticum chuanxiong, Lipinski maxim, Lippia grandifolia, Lippia grandis, Lippia graveolens, Lippia microphylla, Litsea japonica, Lonicera japonica, Lotus japonica, Lotus japonicus, Lumnitzera racemosa, Lutzomyia phlebotomine, Luxemburgia nobilis, Lycopersicon esculentum, Lycopersicum esculentum, Machilus japonica, Machilus thunb, Macleaya cordata, Macleya cordata, Magnaporthe grisea, Magnaporthe oryzae, Magydaris tomentosa, Malbranchea aurantiaca, Mallotus japonicus, Mandragora officinarum, Markhamia tomentosa, Marsdenia tomentosa, Matayba arborescens, Maytenus robusta, Maytenus spinosa, Mentha microphylla, Metaplexis japonica, Miconia willd, Microcos paniculata, Micromonospora aurantiaca, Micromonospora terminalia, Mikania cordifolia, Milbraedia paniculata, Monotes engl, Montanoa tomentosa, Morinda angustifolia, Morinda lucida, Muricidae molluscs, Murraya microphylla, Murraya paniculata, Myrmicaria ants, Myrtus communis, Myxococcus xanthus, Myzus pers, Nectandra grandiflora, Neff trophozoites, Nelumbo nucifera, Nemania diffusa, Nerium indicum, Nicotiana benth, Nigrospora oryzae, Ocotea macrophylla, Olearia paniculata, Onosma paniculata, Ophiopogon japonicus, Ophipogon japonicus, Oroxylum indicum, Ostryopsis nobilis, Ovis orientalis, Padina arborescens, Paenibacillus larvae, Palicourea angustifolia, Panax japonicus, Parasitic helminths, Parastrephia lucida, Parrotia pers, Paulownia tomentosa, Penicillium maxim, Pentaclethra macrophylla, Perilla frutescens, Peronospora arborescens, Perrottetia arborescens, Persicaria odorata, Petasites japonicus, Phaeosphaeria oryzae, Phodophyllum peltatum, Phragmites communis, Physalia physalis, Phytophthora capsici, Phytophythora capsici, Phytoseiulus pers, Picea abies, Pilosella officinarum, Pimenta racemosa, Piper peltatum, Planktothirx rubescens, Planktothrix rubescens, Platanus orientalis, Platycaldus orientalis, Platycladus orientalis, Pleurospermum benth, Pleurotus ferula, Pluchea odorata, Plus pers, Podophyllum peltatum, Podophyllun peltatum, Podospora communis, Popillia japonica, Poria cocos, Portulaca oleracea, Pourthiaea lucida, Prangos ferula, Primula sieb, Prunus pers, Pseudodistoma arborescens, Pseudolabrus japonica, Pseudomonas aurantiaca, Psorothamnus arborescens, Pupalia lappa, Purified cysts, Pyricularia oryzae, Quercus ilex, Rabdosia rubescens, Raphanus sativus, Rhizoctonia solani, Rhizopus oryzae, Rhodomyrtus tomentosa, Rhopilema esculentum, Ricinus communis, Riziocotinia solani, Rubia cordifolia, Rubus sieb, Ruta graveolens, Saccharina japonica, Saccharopolyspora spinosa, Saccharum officinarum, Saffron crocus, Salpichora diffusa, Salvadora pers, Salvia fruticosa, Salvia microphylla, Salvia tomentosa, Sambucus sieb, Sapium indicum, Sargassum thunb, Sarocladium oryzae, Schistosome cercariae, Sclerotinia sclerotiorum, Sclerotonia sclerotiorum, Scophthalmus maxim, Scrophularia lucida, Seminavis robusta, Sesamum indicum, Shorea robusta, Sida spinosa, Siegesbeckia orientalis, Sinnularia maxim, Sinularia maxim, Sitophilus oryzae, Solanum linn, Solieria robusta, Sophora angustifolia, Sophora

| | | |
|---|---|---|
| | | japonica, Spanish lavender, Spilanthes paniculata, Spiraea japonica, Spirulina maxim, Stelletta maxim, Stemona japonica, Stemphylium solani, Stephania japonica, Stigmatella aurantiaca, Streptomyces spinosa, Streptopelia orientalis, Styrax japonica, Swertia japonica, Swietenia macrophylla, Symbiodinium dinoflagellates, Symplococos paniculata, Symplocos racemosa, Syzygium zeylanicum, Tagetes erecta, Tagetes lucida, Teloxys graveolens, Terminalia paniculata, Tetilla japonica, Thuja orientalis, Thymelaea microphylla, Tilia americana, Tilia cordata, Tinospora cordifolia, Torreya nucifera, Total nauplii, Tragopogon orientalis, Triphophyllum peltatum, Triphyophyllum peltatum, Trypanosomatidae protozoa, Trypanosomatidae protozoans, Tuber indicum, Turnera diffusa, Uncaria tomentosa, Undaria pinnatifida, Uvaria macrophylla, Vepris macrophylla, Vernonia paniculata, Veronica pers, Viguiera robusta, Viola odorata, Viridans streptococci, Vismia orientalis, Vitis thunb, Vitro maxim, Vmax maxim, Walker larvae, Walsura robusta, Woodfordia fruticosa, Xanthomonas oryzae, Xenopus frogs, Xenopus laevis, Ximenia americana, Xinjiang provinces, Yucatan wetlands, Zanthoxylum arborescens, Zanthoxylum macrophylla, Zephyranthes robusta, Ziziphus lotus |
| microorganisms species | 2-gram | Alternaria solani, Candida albicans, Candida glabrata, Candida spp, Candida tropicalis, Enterobacter aerogenes, Fusarium solani, Klebsiella pneumonia, Magnaporthe grisea, Phytophthora capsici, Proteus spp, Pyricularia oryzae, Pythium ultimum, Ralstonia solanacearum, Staphylococcus aureus, Streptococcus mutans, Xanthomonas oryzae |
| experimental methods | 1-gram | sonicated, homogenized, incubated, agitated, grinding, subculturing, trypsinization, ultracentrifugation, rinsed, washed, diluted, macerated, cleaned, vortexing, blended, aspiration, pipetting, centrifuged, scraping, decantation, shaken, homogenised, homogenization, ultrasonication, refluxed, electroporation, precultured, trituration, hydrodistillation, percolated, vortexed, lyophilization, charring, fractioned, boiled, autoclaving, kept, resuspended, lysed, reincubated, infused, evaporated, extracted, stirred, stored, preincubated, immersed, percolation, decanted, crystallisation, soaking, streaking, prepared, dissolved, ultrasonicated, thawedflushing, maintained, equilibrated, desalted, passaged, triturated, filtered, removed, placed, grown, desorbed, harvested, coevaporated, decapitation, acidified, discharged, mr., subcultured, immersing, soaked, heated, imaged, centrifuging, titrated, electroblotting, partitioned, poured, autoradiography, redissolved, suspended, solubilized, spun, sown, resuspending, neutralised, degassed, inoculating, drained, spiking, dripped, filtrated, reconstituted, solubilised, centrifugated, coincubated, passaging, azeotroped, perfused, rinsing, collected, suspending, incubating, fractionated, cultured, pelleted, dispersed, planted, repurified, separated, cultivated, cryopreserved, replenished, neutralized, bated, autoclaved, held, quenched, aliquoted, concentrated, smashed, moore, solidification, suction, pipetted, washing, scraped, fixed, schreiber, seeded, plated, swelled, stocked, chromatographed, swabbed, stripped, cleared, desiccated, dried, rotated, rubbing, retested, venipuncture, defatting, profiled, rechromatographed |

[1] The appearance of these n-grams in sentences not necessarily will lead to their extraction as a categorical parameter, because we used other criteria (e.g. high topic correlation, the sentence be a part of the article Experimental section) to confirm that a particular n-gram is used in the context of a determined subject.